\pdfoutput=1
\documentclass[11pt]{article}

\usepackage{EMNLP2022}

\usepackage{times}
\usepackage{latexsym}

\usepackage[T1]{fontenc}
\usepackage[utf8]{inputenc}

\usepackage{microtype}

\usepackage{multirow}
\usepackage{graphicx}
\usepackage{booktabs}
\usepackage{algorithm}
\usepackage{algorithmic}
\usepackage{colortbl}
\usepackage{amsmath}
\usepackage{amssymb}
\usepackage{soul, color, xcolor}
\usepackage{cite}

\usepackage{inconsolata}
\usepackage{cleveref}

\title{Anticipating the Unseen Discrepancy for Vision and Language Navigation}

\author{Yujie Lu* \\
  UC Santa Barbara \\\And
  Huiliang Zhang* \\
  McGill University \\\And
  Ping Nie\\
  Peking University \\\AND
  Weixi Feng\\
  UC Santa Barbara \\\And
  Wenda Xu\\
  UC Santa Barbara \\\And
  Xin Wang\\
  UC Santa Cruz \\\AND
  William Wang \\
  UC Santa Barbara}

\newcommand{\methodname}[1]{DAVIS}
\newcommand{\NNC}[1]{NNC}
\newcommand{\TTA}[1]{TTA}
\newcommand{\clip}[1]{CLIP-ViL}
\newcommand{\bert}[1]{VLN$\protect\circlearrowright$BERT}
\begin{document}
\maketitle

\begin{abstract}
Vision-Language Navigation requires the agent to follow natural language instructions to reach a specific target.
The large discrepancy between seen and unseen environments makes it challenging for the agent to generalize well.
Previous studies propose data augmentation methods to mitigate the data bias explicitly or implicitly and provide improvements in generalization.
However, they try to memorize augmented trajectories and ignore the distribution shifts under unseen environments at test time. 
In this paper, we propose an Unseen \textbf{D}iscrepancy \textbf{A}nticipating \textbf{VIS}ion and Language Navigation (\methodname~) that learns to generalize to unseen environments via encouraging test-time visual consistency.
Specifically, we devise: 1) a semi-supervised framework \methodname~ that leverages visual consistency signals across similar semantic observations. 2) a two-stage learning procedure that encourages adaptation to test-time distribution.
The framework enhances the basic mixture of imitation and reinforcement learning with Momentum Contrast to encourage stable decision-making on similar observations under a joint training stage and a test-time adaptation stage.
Extensive experiments show that \methodname~ achieves model-agnostic improvement over previous state-of-the-art VLN baselines on R2R and RxR benchmarks.\footnote{Our source code and data are in supplemental materials.}

\end{abstract}

\section{Introduction}
Vision-and-language navigation (VLN) tasks have attracted increasing research interests and achieved significant improvements with the emergence of deep learning techniques.
VLN is a complex system that requires decision-making conditioned on visually grounded language understanding.
There are some unique challenges in the VLN task, such as reasoning over cross-modal input and generalizing to unseen environments.

Previous studies~\citep{Wang2019reinforcedvln, Tan2019LearningTN, shen2021much, Hong2021VLNBERTAR} addressed these issues by proposing to enforce cross-modal grounding via imitation learning and reinforcement learning.
To improve the generalizability, most prior studies~\citep{fu2020counterfactual, Majumdar2020ImprovingVN, Liu2021VisionLanguageNW, Parvaneh2020CounterfactualVN} propose diversified forms of augmentations on input data.
Other studies~\citep{Li2019RobustNW, Zhu2020VisionLanguageNW} explore the utilization of self-supervision from the data or prior knowledge from the pre-trained models.
These studies mainly focused on designing techniques that respect training time generalization.
However, due to the distribution shift at test time, these studies suffer from the difficulty of generalizing and maintaining robustness under this situation.
For instance, test-time decision-making should be invariant to the changes of irrelevant objects in the environment and the changes of the viewpoint.
The robustness of vision-and-language agents against these test-time shifts requires more research focus in this area.
\begin{figure*}[t]
    \centering
    \includegraphics[width=.95\textwidth]{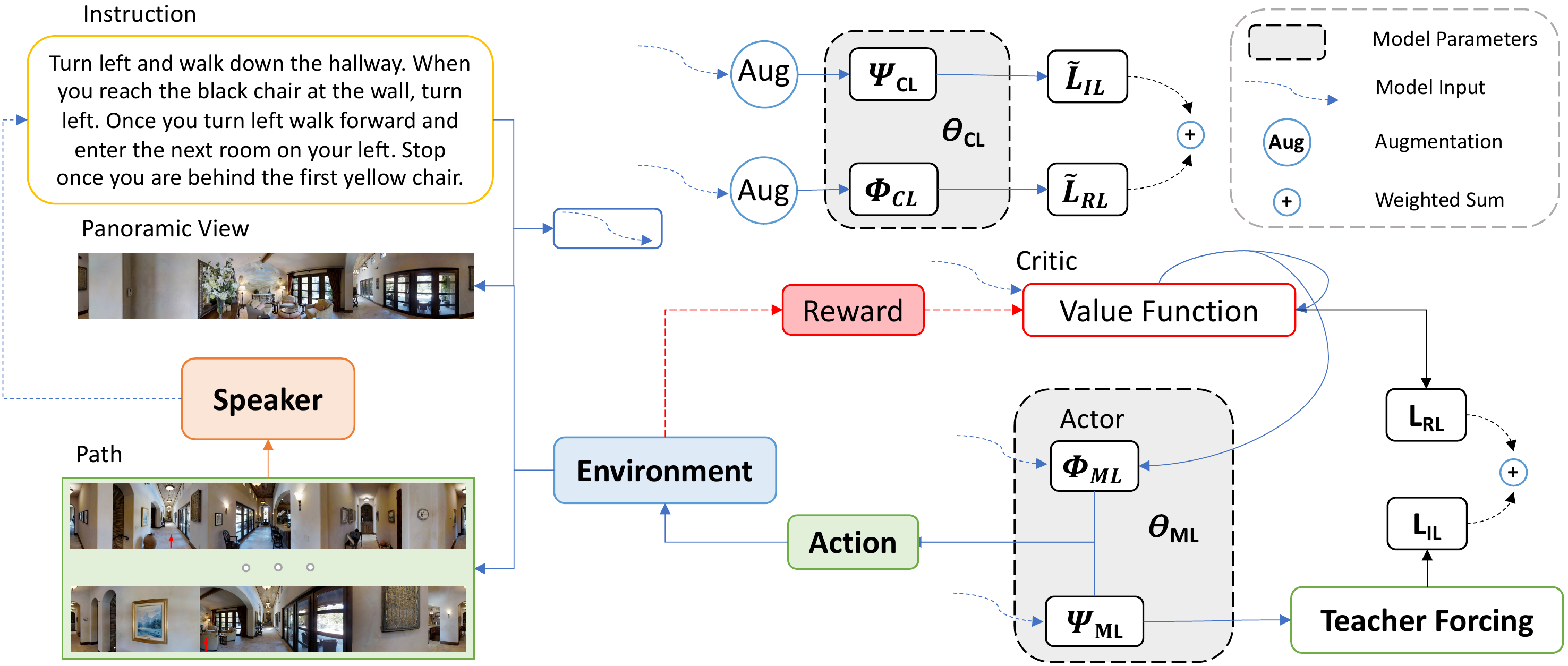}
    \caption{\textbf{Overall Architecture.}
    \methodname~ consists of supervised $\theta_{ML}$ and self-supervised $\theta_{CL}$.
    $\psi$ and $\phi$ provide Imitation Learning $\mathcal{L}_{IL}$ and Reinforcement Learning $\mathcal{L}_{RL}$ objectives, respectively.
    }
    \label{fig:overview}
\end{figure*}

In this paper, we study test-time visual consistency (NNCVLN) in the context of VLN. We also demonstrate that a robustly adapted VLN agent at test-time with self-supervision can outperform prior approaches, which suffer from the biased  training data distribution.
We propose a test-time robust vision-and-language navigation framework, which consists of self-supervised and supervised modules for both imitation and reinforcement learning.
We adopt two general architectures for the supervised module, sequence-to-sequence, and transformer-based, which can translate human instructions grounded in the visual environment to history contextualized action sequences.
For the self-supervised module, we utilize contrastive learning to encourage the agent to learn the visual consistency and invariant features from the observations.
We apply instance-level augmentation methods for visual environment input to generate positive samples.
To obtain negative samples, we select the temporal different visual input from the same path.
Such consistency is then learned via encouraging the agent to predict similar action distributions over positive instruction-observation pairs and dissimilar ones otherwise.
To ensure the agent adapted to the distribution shift, we devise the two-stage training strategy.
First, the agent is trained with the semi-supervised objectives by encouraging the predictions to obtain minimized entropy across augmentations over train data.
Second, the agent utilizes the self-supervised objectives by encouraging minimized entropy across augmentations over test data and updating the framework before inference.

Our Test-time Visual Consistency (\methodname~) framework is a general learning paradigm that can be easily applied to existing VLN baselines to boost their performance of generalization to new scenes.
Extensive experiments show that \methodname~ achieves model-agnostic 
improvement over previous state-of-the-art VLN baselines on R2R~\citep{Anderson2018VisionandLanguageNI} and RxR~\citep{ku2020room} Benchmarks.
In summary, our contributions are three-fold:
\begin{itemize}
    \item We propose a semi-supervised VLN framework \methodname~ that enforces stable decision-making via visual consistency regularization.
    \item To improve generalizability to unseen environments and maintain robustness under distribution shift, we devise a two-stage training strategy for the VLN agent, consisting of semi-supervised training and self-supervised test-time adaptation.
    \item Empirically, the model-agnostic \methodname~ achieves consistent performance gain over state-of-the-art VLN baselines disregarding the architectures over R2R and RxR Benchmarks.
\end{itemize}

\section{Methodology}
\label{sec:method}

\subsection{Problem Definition}
The vision-language navigation task requires an agent to follow the instruction ${I}$ in a photo-realistic environment ${E}$.
% \hl{1 do we need to omit the number 36 here, just illustrate a general problem definition? 2 Or illustrate more on the VLN task? 3 or what instruction is and policy is ?}
The panoramic view ${o_t}$, which is divided into ${36}$ single views ${\{o_{t,i}\}_{i=1}^{36}}$, are provided at each time step $t$.
The agent start from the viewpoint ${S}$ and make actions decisions by selecting from the navigable viewpoints with the policy network $\pi$.  
After navigating to the new viewpoint, the agent observes the new panoramic view.
Finally, the agent stop at the predicted target position $T$ and give the navigation route $R$.
% The navigable viewpoints ${\{l_{t,k}\}_{k=1}^{N_{t}}}$ are the ${N_t}$ reachable and visible locations from the current viewpoint. Each navigable viewpoint ${l_{t,k}}$ is represented by the orientation ${(\hat{\theta_{t,k}}, \hat{\phi_{t,k}})}$ from current viewpoint to the next viewpoints. The agent needs to select the moving action ${a_t}$ from the list of navigable viewpoints ${l_{t,k}}$ according to the given instruction ${I}$, panoramic views ${\{o_{\delta}\}_{\delta=1}^{t}}$, and action sequences ${\{a_\delta\}_{\delta=1}^{1:t-1}}$.

\subsection{Overall Architecture}
The overall architecture of \methodname~ is shown in Figure~\ref{fig:overview}.
Given the instruction $I$ and panoramic view ${o_t}$ extracted from photo-realistic environment ${E}$ at time step $t$, the navigator $N$ predicts action $a_t$, which is applied to $E$ to generate trajectory path $P$. 
% \hl{1 Do we need to draw  the navigator in Fig? 2 or illustrate the I E N in fig? 3 draw the two-stage in fig? 4 or emphasis the  two parts in Fig 1? yes, need to revise}
The speaker $S$ learns to translate the $P$ to $\hat{I}$ maximally similar to $I$.
The proposed \methodname~ is composed of two parts: 1) Supervised part $\psi_{ML}$ (IL) and $\phi_{ML}$ (RL). 2) Self-supervised part $\psi_{CL}$ and $\phi_{CL}$.
% supervised
$\psi_{ML}$ learns to imitate the teacher actions $\hat{a}_{t}$ at each time step $t$.
$\phi_{ML}$ learns to maximize the reward return through the trajectory path.
% self-supervised
$\psi_{CL}$ learns to maximize the similarity of the actions across similar visual observation at each time step $t$.
$\phi_{CL}$ learns to maximize the similarity of reward return through similar trajectory paths.
We devise a two-stage training strategy for \methodname~: 1) At the training stage, both supervised and self-supervised parts are jointly trained. 2) At test-time, the supervised part is fixed, while the self-supervised continue to adapt.

\subsection{Model}
\label{sec:model}
\noindent \textbf{Grounded Navigation Reasoning}
Our model design is agnostic to different architectures of navigation reasoning grounded in language and vision understanding.
Hence, we select two typical navigation reasoning architectures that conditioned on natural language instruction and visual environment observation as our baselines, \textit{a)} Sequence to Sequence. \textit{b)} Transformer-based.
For sequence-to-sequence architecture, the encoder-decoder model with a bi-directional LSTM-RNN encoder and an attentive LSTM-RNN decoder are adopted following EnvDrop~\citep{Tan2019LearningTN}.
The instruction-to-navigation translation is computed, first by calculating the attentive visual feature at each decoding step $t$ as:
\begin{equation}
    \hat{f}_t = \sum_{i}\text{softmax}_{i}(f^{T}_{t,i} W_{F} \hat{h}_{t-1}) f_{t,i}
\end{equation}

The instruction-aware hidden output $\hat{h_{t}}$ is computed with the combination of hidden output of the LSTM and the attentive instruction feature.
\begin{equation}
    \hat{h}_{t} = tanh(W[\hat{u}_{t}; LSTM([\hat{f}_t; \hat{a}_{t-1}], \hat{h}_{t-1})])
\end{equation}
where $\hat{u}_{t}$ is the attentive instruction feature, $\hat{a}_{t-1}$ is the previous action.
For transformer-based architecture, we follow \bert~ to process the language.
We get initial state token $s_{0}$ and tokens $L^{I}=u_{i}$ for instruction $I={w_{i}}$ by:
\begin{equation}
    s_{0}, L^{I} = E_{language}(\texttt{[CLS]}, I, \texttt{[SEP]})
\end{equation}
where $E_{language}$ is the language encoder of the BERT. $\texttt{[CLS]}$ and $\texttt{[SEP]}$ are predefined tokens in the BERT model.
Here, we use $\texttt{[CLS]}$ to represent the initial state.
The visual tokens $V_t$ will refer to the language token $L$ as keys and values for observation and instruction attention at each navigation step $t$.
For each raw observation $o^{raw}_t$ at time step $t$, it is projected to the textual space of the BERT embedding.
And then the final input $X_t$ at the current time step is the concatenation of the state token $s_t$, textual and visual tokens:
\begin{equation}
    V^{o}_t = I_{t} W_{I}, X_{t} = [s_t, L^{I}_t, V^{o}_{t}]
\end{equation}
Then the state will be represented as the summary of both textual and visual tokens of all the previous observations, history action sequences.
The raw textual and visual tokens are matched with state and language attention weights, which is computed by averaging the scores from $K$ attention heads as:
\begin{equation}
    \hat{A_{l}} = \text{softmax}(\hat{A}_{l}) = \text{softmax}(\frac{1}{K}\sum^{K}_{k=1} \frac{Q_{l,k} K_{l,k}}{\sqrt{d_{h}}})
\end{equation}
where $Q_{l,k}$ and $K_{l,k}$ represent the state tokens as query matrix and instruction tokens as the key matrix at head $k$ of final layer $l$.

\begin{algorithm}[t]
\caption{Test-time Visual Consistency}
\label{algo:procedure}
  \algsetup{linenosize=\small}
  \small
\begin{algorithmic}[1]

\REQUIRE ~~\\ %Input
Train dataset $\mathcal{D}_{train}$, validation dataset $\mathcal{D}_{val}$ and test dataset ${\mathcal{D}_{test}}$;\\
Supervised module parameters $\theta_{ML}$ consists of $\phi_{ML}$ and $\psi_{ML}$; \\
Self-supervised consistency module parameters $\theta_{CL}$ consists of $\phi_{CL}$ and $\psi_{CL}$;\\
% Model learning rate $\epsilon_{1}$ for training stage and $\epsilon_{2}$ for testing stage;\\
Augmentation Function Pool $\mathcal{A}$;\\
\ENSURE ~~\\ %Output

\FOR{each iteration}
\FOR{each N in $\mathcal{D}_{train}$}
\STATE Sample augmentation function $F_{a}$ from Pool $\mathcal{A}$, apply Augmentation $F_{a}$, $o' = f(o)$;
\STATE Compute loss
\STATE $\theta^{i}_{ML}$ = $\theta^{i-1}_{ML}$ + update, $\theta^{i}_{CL}$ = $\theta^{i-1}_{CL}$ + update
\ENDFOR
\ENDFOR

\FOR{each test sample in ${\mathcal{D}_{test}}$ }
\STATE Sample augmentation function $F_{a}$ from Pool $\mathcal{A}$, apply Augmentation $F_{a}$, $o' = f(o)$;
\STATE Compute loss
\STATE $\theta^{i}_{CL}$ = $\theta^{i-1}_{CL}$ + update
\ENDFOR

\STATE Inference;
\end{algorithmic}
\end{algorithm}

\noindent \textbf{Data Augmentation}
Navigation in the unseen environment requires a model learning invariances across diversified observations.
For example, the viewpoint of the camera and the irrelevant objects changes should not influence the agent's decision-making process.
Data augmentations are usually utilized to encourage the model to encode the invariant features to achieve such generalization.
To guarantee that the model generalizes at test time under these unseen invariances, we augment the sample at test time and encourage the model to respect invariances and thus maintain robustness even with distribution shift.
For each step $t$, we sample an augmentation function $F_{a}$ from the function pool $A = \{F_{1}, F_{2}, ..., F_{N}\}$ and apply it over the visual observation.
\begin{equation}
    \hat{O}^{a}_{t} = F_{a}(O^{raw}_{t}), a \in [1, N], F_{a} \in A
\end{equation}
where $O^{raw}_{t}$ is the raw panoramic features obtained from the environment. $\hat{O}^{a}_{t}$ is the augmented observations at time step $t$ with augmentation function $F_{a}$.
For the navigation domain, we devise the instance-level visual modifications following EnvDrop~\citep{Tan2019LearningTN} and feature-level modifications by applying the dropout layer. 
After applying augmentations over the samples, the model takes the augmented samples for further self-supervised module and adapt them before inference.

% \begin{figure*}[t]
%     \centering
%     \includegraphics[width=\textwidth]{figures/contrastive.pdf}
%     \caption{\textbf{Contrastive Module} (a) The policy $\psi$ learns by attracting positive samples in the predicted action embedding space. (b) The policy $\phi$ regularizes the reward for being similar across positive samples ($a_{t}$ and $\dot{a}_{t}$) for RL. The visual consistency is encouraged between visual embedding of Encoder and Momentum Encoder.}
%     \label{fig:vlnContrastive}
% \end{figure*}
% \noindent \textbf{Visual Consistency}
\noindent \textbf{Consistent Semantic Observation}
The visual encoder is the basic component of the VLN agent, on which the speaker relies for grounding the instructions and the navigator relies on the decision making.
To ensure that the encoded observation obtains the consistent semantic meaning, we train the encoders by applying the similarity dot product, which measures agreement between raw observation (query) and augmented observation (key) pair.
We follow MoCo~\citep{He2020MomentumCF} to use the averaged momentum of the query encoder to encode the augmented views in the key queue as a momentum encoder.
The parameters $\theta_{k}$ of the momentum encoder is updated as:
\begin{equation}
    \theta_{k} \leftarrow m\theta_{k} + (1-m)\theta_{q}, m \in [0, 1)]
\end{equation}
where $\theta_{q}$ is the parameters of the query encoder that is updated by back-propagation, and  $m$ is a momentum coefficient.

\noindent \textbf{Consistent Teacher Forcing}
% \hl{Do we need to combine 3.3.4 and 3.3.5 and 3.3.6?}
Typically, we have the Teacher Forcing technique to translate with a supervised Imitation Learning Module from instructions to action sequences.
Specifically, the agent navigates on the ground-truth trajectory by following teacher actions and calculates a cross-entropy loss for each decision.
Furthermore, we consider taking augmentations for introducing consistency during teacher forcing.
% As shown in Figure~\ref{fig:vlnContrastive}(a), 
The policy $\psi_{CL}$ network learn to maximize the agreement of action decisions $a_t$ and $\hat{a}_t$ over positive observation pairs.

\begin{figure}[t]
    \centering
    \includegraphics[width=0.9\columnwidth]{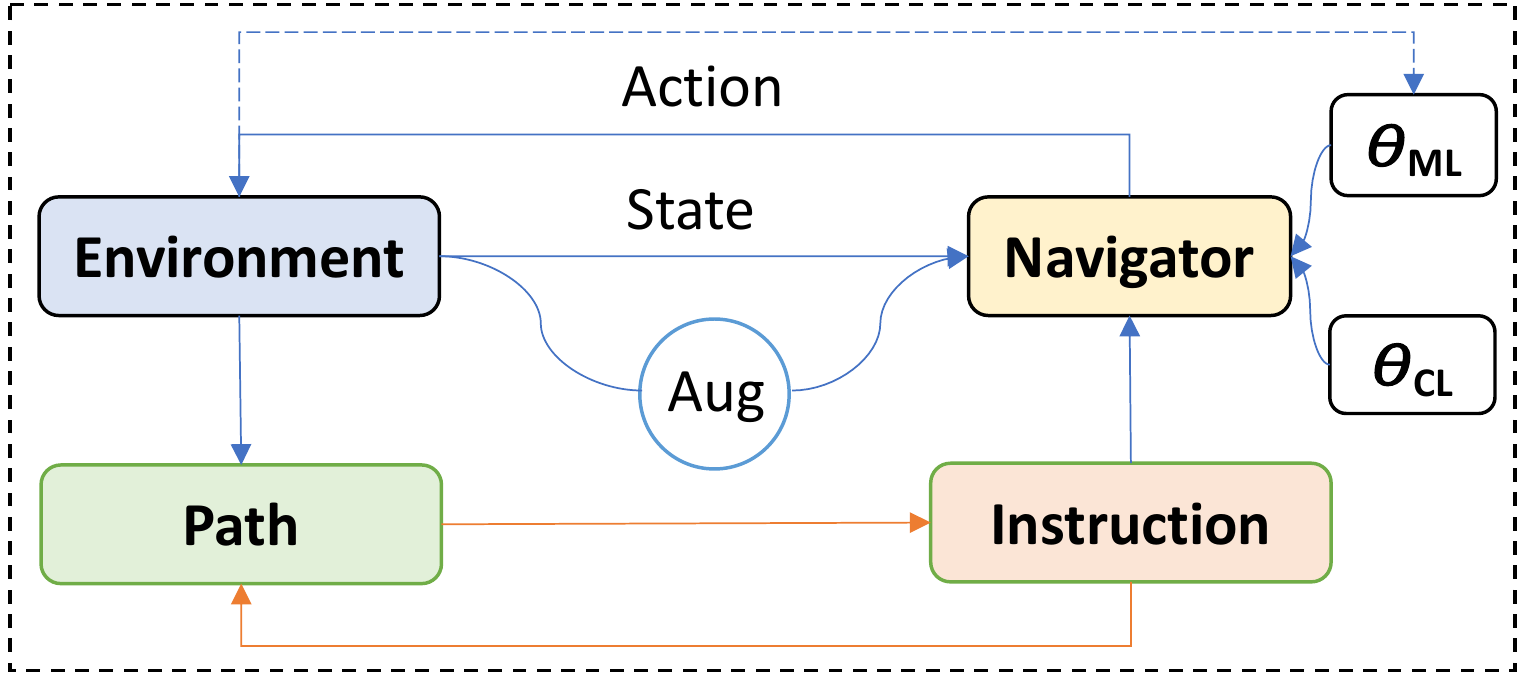}
    \caption{\textbf{Training and Test-time Procedure.} For the training stage, the $\theta_{ML}$ and $\theta_{CL}$ are updated jointly with Equation~\ref{eq:obtrain}. At test-time, the $\theta_{ML}$ is fixed, and the $\theta_{CL}$ is adapted using Equation~\ref{eq:obtest}.
    % After the adaptation, the action distribution is predicted by the combination of $\theta_{ML}$ and $\theta_{CL}$.\yj{TODO: update description}
    }
    \label{fig:procedure}
\end{figure}

\noindent \textbf{Consistent Soft Actor Critic}
Typically we have a reinforcement learning method to roll out and update the policy model given the ground truth state comparison as a supervised Reinforcement Learning Module.
Specifically, we apply soft actor critic~\citep{Haarnoja2018SoftAO}, an off-policy RL algorithm that optimizes a stochastic policy for maximizing the expected trajectory returns.
% SAC learns a policy $\pi_{\phi}$ and critics $Q_{\phi}$.
% As shown in Figure~\ref{fig:vlnContrastive}(b), 
We consider taking augmentations for introducing consistency during actor critic learning.
Similar to the query-key encoder learning in visual encoder, we learn the momentum update for the key critic $\phi_{k}$:
\begin{equation}
    \phi_{k} \leftarrow m\phi_{k} + (1-m)\phi_{q}, m \in [0, 1)]
\end{equation}
where $\phi_{q}$ is the parameters of the query critic and updated by back-propagation,  $m$ is a momentum coefficient.
The learning objective is described in Section~\ref{sec:learning}.

\noindent \textbf{Action Prediction}
The output embeddings ${z^v_{1:T}}$ go through a single fully-connected layer to predict agent actions ${\hat{a}_{1:T}}$. During training, the input is the sequence of observation-language paired data and ground truth action sequence. During testing at timestep ${t}$, we input visual observations ${v_{1:t}}$ and previous actions ${\hat{a}_{1:t-1}}$ taken by the agent.
After model adaptation during test-time, we select the action predicted for the last time step ${\hat{a}_t}$. And then apply ${\hat{a}_t}$ to the environment which generates the next visual observation ${v_{t+1}}$.
Specifically, for sequence-to-sequence based architecture, the decision will be made by $p_t(a_{t,k}) = \text{softmax}_{k}(g^{T}_{t,k}W_{G}\hat{h_{t}})$.
For transformer-based architecture, the decision will be made by $p^{a}_{t} = \hat{A}^{s,v}_{l}$.

\subsection{Learning}
\label{sec:learning}
We learn a generalizable policy that handles the distribution shift between train and test data.
The learning procedure is devised in two phase:1) Semi-supervised learning on train data split. 2) Self-supervised adaptation on test-time data.
$\theta_{CL}$ and $\theta_{RL}$ is jointly learned with the full objectives aggregated by $\tilde{\mathcal{L}_{IL}}$, $\tilde{\mathcal{L}_{RL}}$, $\mathcal{L}_{IL}$ and $\mathcal{L}_{RL}$ at standard training stage.
At test-time adaptation, $\theta_{ML}$ is fixed while $\theta_{CL}$ is updated before inference.
The learning procedure is summarized in Algorithm~\ref{algo:procedure}.

\noindent \textbf{Standard Training}
The standard training procedure is shown in Figure~\ref{fig:procedure}.
For the imitation learning, the objective is as:
\begin{equation}
    \mathcal{L}^{IL} = \sum_{t} \mathcal{L}^{IL}_{t} = \sum_{t} -a^{*}_{t}\log p_{t}(a_{t})
\end{equation}
% Reward function:?
For the reinforcement learning, the training objective is:
\begin{equation}
    \mathcal{L}^{RL}(\phi, B) = E_{t~B} [(Q_{\phi}(o, a) - (r + \gamma(1-d)T))^{2}]
\end{equation}
where $t=(o,a,o',r,d)$ is a tuple with observation $o$, action $a$, reward $r$ and done signal $d$, $B$ is the replay buffer, and $T$ is the target, defined as:
\begin{equation}
    T = (\min Q^{*}_{\phi} (o', a') - \alpha \log \pi_{\phi} (a'|o'))
\end{equation}
where $Q^{*}_{\phi}$ is the exponential moving average of the parameters of $Q_{\phi}$ to improve training stability in off-policy RL algorithms.
$\alpha$ determines the priority of the entropy maximization over value function optimization.
The critic is trained by maximizing the expected return of its actions as:
\begin{equation}
    \mathcal{L}(\phi) = E_{a~\pi}[Q^{\pi}(o, a) - \alpha\log\pi_{\phi}(a|o)]
\end{equation}
where actions are sampled stochastically from the policy.
Then we aggregate these two objectives as our supervised objective $\mathcal{L}^{ML}$:
\begin{equation}
    \mathcal{L}^{ML} = \mathcal{L}^{RL} + \lambda \mathcal{L}^{IL}
\end{equation}
where $\lambda$ balances the weight of Imitation Learning objective and Reinforcement Learning objective.

To leverage the information from data, we add self-supervised objectives during navigation.
The contrastive teacher-forcing module learns with the objective:
\begin{equation}
    \mathcal{L}^{IL}_{CL} = \log \frac{exp(q^{T}Wk_{+})}{exp(q^{T}Wk_{+}) + \sum_{i=0}^{K-1}exp(q^{T}Wk_{i})}
\end{equation}
The contrastive soft actor critic module learns with the objective:
\begin{equation}
    \mathcal{L}^{RL}_{CL} = \log \frac{exp(q^{T}Wk_{+})}{exp(q^{T}Wk_{+}) + \sum_{i=0}^{K-1}exp(q^{T}Wk_{i})}
\end{equation}
Please refer to Appendix~\ref{sec:method_details} for the details of the contrastive module.

Finally, we aggregate above two objectives for self-supervised training objective:
\begin{equation}
    \mathcal{L} = \mathcal{L}^{ML} + \lambda \mathcal{L}^{CL}
\label{eq:obtrain}
\end{equation}

\noindent \textbf{Test-time Adaptation}
The test-time procedure is shown in Figure~\ref{fig:procedure} with separate detailed description in Figure~\ref{fig:procedure_all}(b).
At test time, we remove the supervised objective and adapt the model with only the self-supervise objective on test-time sample as:
\begin{equation}
    \mathcal{L}_{CE}(\theta;x) = \frac{1}{B} \sum_{i=1}^{B}H(p_{\theta}(\dot | \hat{x_{i}}))
\label{eq:obtest}
\end{equation}

\section{Experimental Setup}
\label{sec:exp_setup}
\begin{table*}[t]
\centering
\resizebox{.9\textwidth}{!}{%    
\begin{tabular}{l ll llll llll}
\toprule
& & \multicolumn{4}{c}{\textbf{Validation Seen}} & \multicolumn{4}{c}{\textbf{Validation Unseen}} \\
\cmidrule(lr){3-6}\cmidrule(lr){7-10}\
&\textbf{Model} & TL & NE $\downarrow$ & SR $\uparrow$ & SPL $\uparrow$ & TL & NE $\downarrow$ & SR $\uparrow$ & SPL $\uparrow$ \\
    \midrule
    \texttt{0}&Random     & 9.58 & 9.45 & 16 & - & 9.77&9.23 &16  &- \\
    \texttt{1}&Human      &- &- &- &- &- &- &-  &- \\
    \midrule
    \texttt{2}&Seq2Seq~\citep{Anderson2018VisionandLanguageNI}         &11.33 &6.01 & 39&- &8.39 &7.81 &22  &- \\
    \texttt{3}&Speaker-Follower~\citep{fried2018speakerfollower}          & -&3.36 &66 &- &- &6.62 &35  &-\\
    \texttt{4}&PRESS~\citep{Li2019press}        & 10.57 &4.39 &58 &55 &10.36 &5.28 &49  &45\\
    \texttt{5}&AuxRN~\citep{Zhu2020AuxRN}        & - &3.33 &70 &67 &- &5.28 &55  &50 & \\
    \texttt{6}&PREVALENT~\citep{Hao2020prevalent}        & 10.32&3.67 &69 &65 &10.19 &4.71 &58  &53 \\
    \texttt{7}&RelGraph~\citep{Hong2020relgraph}        &10.13 &3.47 &67 &65 &9.99 &4.73 &57  &53 \\
    \midrule
    \texttt{8}&\clip~~\citep{shen2021much}      & 12.59 &4.13 &60.32 &55.44 & 11.46& 5.13 & 52.45 &46.76 \\
    \rowcolor{gray!10} \texttt{9}&~~~ $+$ Nearest Neighbor Contrastive        &9.97 &4.40 &63.51 &57.58 &9.62 & 4.45& 52.19 & 48.24 \\
    \rowcolor{gray!10} \texttt{10}&~~~ $+$ Test-time Adaptation (~\methodname~)       &10.16 &4.05 &64.67 &61.50 &10.46 & 4.41 & 54.04 & 49.14\\
    \texttt{11}&\bert~~\citep{Hong2021VLNBERTAR} & 10.92 & \underline{2.96} & 72.48 &68.03  & 11.04 & 4.13 & 59.86 & 55.09\\
    \rowcolor{gray!10} \texttt{12}&~~~ $+$ Nearest Neighbor Contrastive    & 11.13 & \textbf{2.66} & \underline{74.53} & \underline{69.92} & 11.04 & \underline{3.92} & \underline{62.22} & \underline{56.56}\\
    \rowcolor{gray!10} \texttt{13}&~~~ $+$ Test-time Adaptation (~\methodname~)     & 12.45 & 3.16 & \textbf{80.48} & \textbf{76.19} & 12.65 & \textbf{3.16} & \textbf{67.18} & \textbf{60.51} \\
    \bottomrule
\end{tabular}
}
    \caption{\textbf{Model-agnostic Improvement over Powerfull Baseliens on R2R Benchmark.} \methodname~ is our full architecture applied over baselines. \textbf{BEST} and the \underline{SECOND} best results are highlighted.
    }
    % \vspace{-4ex}
    \label{tab:r2r}
\end{table*}

\begin{table*}[t]
\centering
\resizebox{.9\textwidth}{!}{%    
\begin{tabular}{l l lllll lllll}
\toprule
& &\multicolumn{5}{c}{\textbf{Validation Seen}} & \multicolumn{5}{c}{\textbf{Validation Unseen}} \\
\cmidrule(lr){3-7}\cmidrule(lr){8-12}\
&\textbf{Model} & SR $\uparrow$ & SPL $\uparrow$ & CLS $\uparrow$ & nDTW $\uparrow$ & sDTW $\uparrow$ & SR $\uparrow$ & SPL $\uparrow$ & CLS $\uparrow$ & nDTW $\uparrow$ & sDTW $\uparrow$ \\
    \midrule
    \texttt{1}&EnvDrop~\citep{Tan2019LearningTN}         & - & - & - & - & - & 38.5& 34& 54& 51& 32\\
    \texttt{2}&Syntax~\citep{Li2021syntax}           & - & - & - & - & - & 39.2& 35& 56& 52& 32\\
    \midrule
    \texttt{3}&\clip~~\citep{shen2021much}        & 44.35 & 41.39 & 59.96 & 56.43 & 38.41 & 39.72& 35.66 & 55.35 & 51.85 & 32.51\\
    \rowcolor{gray!10} \texttt{4}&~~~ $+$ Nearest Neighbor Contrastive  & 46.18 & 45.43 & 61.18 & 57.98 & 41.06 & 40.11& 36.45 & 56.73 & 53.08 & 33.73\\
    \rowcolor{gray!10} \texttt{5}&~~~ $+$ Test-time Adaptation (~\methodname~)  & 47.72&46.71 & 61.80&58.30 &  42.00&40.72& 38.32&\underline{57.80} &53.80 & 34.10 \\
    \texttt{6}&\bert~~\citep{Hong2021VLNBERTAR}     & 48.21 & 45.84&61.32 &59.23 & 41.86& 43.16 &38.32 &56.24 & 53.12& 35.21\\
    \rowcolor{gray!10} \texttt{7}&~~~ $+$ Nearest Neighbor Contrastive        & \underline{49.50}& \underline{47.12} &\underline{61.90} & \underline{60.01} &\underline{42.53} & \underline{46.01} &\underline{39.09} & 57.33& \underline{54.11} & \underline{36.80}\\
    \rowcolor{gray!10} \texttt{8}&~~~ $+$ Test-time Adaptation (~\methodname~)  & \textbf{50.64} & \textbf{48.33} & \textbf{62.05} & \textbf{61.32}& \textbf{43.80} &\textbf{47.66} &\textbf{39.85} &\textbf{58.07} & \textbf{54.52}&\textbf{37.16}  \\
    \bottomrule
\end{tabular}
}
    \caption{\textbf{Model-agnostic improvement over baselines on RxR Benchmark.} \methodname~ is our full architecture applied over baselines. \textbf{BEST} and the \underline{SECOND} best results are highlighted.}
    \label{tab:rxr}
\end{table*}
\noindent \textbf{Datasets}
We evaluate our methods on following two benchmarks for vision-language navigation in real 3D environments.
\textbf{R2R}~\citep{Anderson2018VisionandLanguageNI} is a VLN dataset collected in photo-realistic environments (Matterport3D~\citep{Chang2017Matterport3DLF}).
\textbf{RxR}~\citep{ku2020room} is a multilingual (English, Hindi, and Telugu) and larger than other existing VLN datasets with $11089$, $232$, $1517$ and $2684$ paths in train, val-seen, val-unseen, and test splits respectively.
Please refer to Appendix~\ref{sec:dataset_details} for dataset details.

\noindent \textbf{Evaluation Metrics}
On the R2R benchmark, we use standard metrics: Trajectory Length (TL), Navigation Error (NE $\downarrow$), Success Rate (SR $\uparrow$), and Success weighted by inverse Path Length (SPL $\uparrow$). 
On RxR benchmark, we use the standard metrics: SR $\uparrow$, SPL $\uparrow$, Coverage weighted by Length Score (CLS $\uparrow$~\citep{Jain2019StayOT}), Normalized Dynamic Time Warping (nDTW $\uparrow$~\citep{Ilharco2019GeneralEF}), and Success weighted by normalized Dynamic Time Warping (sDTW $\uparrow$~\citep{Ilharco2019GeneralEF}). \\
\noindent \textbf{Baselines}
We compare the single-run performance of different agents (see Table~\ref{tab:r2r} and Table~\ref{tab:rxr}) on R2R and RxR.
We conduct experiments over both benchmarks with our model-agnostic \methodname~ added to baselines: 1) \clip~~\citep{shen2021much} is an extension of EnvDrop~\citep{Tan2019LearningTN}, which leverages the representation from CLIP~\citep{Radford2021LearningTV} to replace the visual features extracting backbone from ResNet~\citep{He2016DeepRL} to Vision Transformer (ViT~\citep{Dosovitskiy2021AnII}). This baseline utilize back-translation as their test-time adaption. 2) \bert~~\citep{Hong2021VLNBERTAR} proposes a recurrent BERT model that is time-aware for use in VLN. The parameters of our implemented baseline are initialized from PREVALENT~\citep{Hao2020prevalent}.
More details of the baselines can be found in Appendix~\ref{sec:baseline_details}.
And please refer to Appendix~\ref{sec:implementation_details} for the experimental implementation details with configuration illustrations.
Note that for each base architecture, we first implement the \textbf{Nearest Neighbor Contrastive} (\NNC~) framework, and then apply the Test-time Adaptation(\TTA~).
\NNC~ is a train-stage version of the \TTA~ which leverage visual consistency from the augmented views as described in \textit{Consistent Semantic Observation} paragraph of Section~\ref{sec:model}.

% \noindent \textbf{Configuration}

\begin{table*}[t]
\centering
\resizebox{.9\textwidth}{!}{%    
\begin{tabular}{l ll lll llll llll llll}
\toprule
& & \multicolumn{3}{c}{\textbf{Losses}} & \multicolumn{4}{c}{\textbf{Validation Seen}} & \multicolumn{4}{c}{\textbf{Validation Unseen}}  \\
\cmidrule(lr){3-5}\cmidrule(lr){6-9}\cmidrule(lr){10-13}\
&\textbf{Model} & ML & $CL_{IL}$ & $CL_{RL}$ & TL & NE $\downarrow$ & SR $\uparrow$ & SPL $\uparrow$ & TL & NE $\downarrow$ & SR $\uparrow$ & SPL $\uparrow$ \\
    \midrule
    \texttt{1}&\clip~ & \checkmark & \checkmark & & 10.66      & 4.33& 63.83& 52.76& 10.15&5.04 &49.25 &44.45 \\
    2&\clip~ & \checkmark & & \checkmark & 10.34          &4.27 &62.15 &50.05 &9.78 &5.08 & 48.75& 46.59\\
    3&\clip~ &  & \checkmark & \checkmark & 11.36 &5.36 &47.48 &42.57 &10.69 &5.65 &43.67 &38.40\\
    \midrule
    4&\bert~  & \checkmark & \checkmark & &  11.55     &\textbf{3.92} &\underline{68.85} & \underline{54.74}& 11.26& \underline{4.93} & \textbf{54.03} & \underline{47.31}\\
    5&\bert~  & \checkmark & & \checkmark & 11.52         & \underline{4.10}& \textbf{69.35} & \textbf{56.46}& 12.37& \textbf{4.56} &\underline{53.35}& \textbf{47.63} & \\
    6&\bert~  &  & \checkmark & \checkmark & 12.70 & 4.20&57.59 &51.89 &13.02 &5.23 &50.75 & 42.99 & \\
    \bottomrule

\end{tabular}
}
    \caption{\textbf{Ablation on R2R Benchmark.} $ML$, $CL_{IL}$ and $CL_{RL}$ represent supervised loss, contrastive loss of imitation learning and reinforcement learning respectively. Full model (\methodname~) in Table~\ref{tab:r2r} consists of all three losses. \textbf{BEST} and the \underline{SECOND} best results are highlighted.}
    \label{tab:r2rablation}
\end{table*}

\section{Experimental Results}
\subsection{Performance Comparison}
We report the evaluation performance of the proposed framework in a model-agnostic setting over powerful baselines, \clip~ and \bert~ under Validation Seen and Validation Unseen splits.
% To validate the model-agnostic effectiveness of our proposed \methodname~, we conduct experiments over both sequence-to-sequence based and transformer-based architectures, as described in Section~\ref{sec:method}.

As shown in Table~\ref{tab:r2r}, our base model initialized with \bert~ performs better than previous methods on the \textbf{R2R} benchmark.
For \clip~, \NNC~ achieve $3.86\%$ and $3.17\%$ improvement compared with \clip~ baseline on SPL over Validation seen and unseen, respectively.
With \TTA~, the model achieves further consistent improvement of $10.93\%$ and $5.09\%$ over baseline accordingly.
This indicates the effectiveness of test-time adaptation.
Similarly for \bert~ baseline, \NNC~ achieves $2.78\%$ and $2.67\%$ improvement, \TTA~ achieves $12.00\%$ and $9.84\%$ improvement on SPL over validation seen and unseen respectively.
In general, for the metrics (TL, NE, SR, and SPL) that research recognizes for the R2R benchmark, our proposed \methodname~ achieve consistent performance gain, which indicates effective generalization to the unseen environment.
Results on the \textbf{RxR} benchmark are provided in Table~\ref{tab:rxr}.
We consider SR, SPL, CLS, nDTW, and sDTW metrics on Validation seen and unseen split.
% \clip~ and \bert~ are powerful baselines on this benchmark.
For \clip~, with \NNC~, the model consistently improves $2.37\%$ and $3.78\%$ on nDTW and sDTW accordingly in Validation unseen.
With \TTA~, the model achieves further consistent improvement of $3.76\%$ and $4.89\%$ accordingly.
We also observe consistent performance gain of \NNC~ over \bert~ with $1.86\%$ and $4.51\%$ on nDTW and sDTW respectively in Validation unseen.
And the model updated with \TTA~ achieves further consistent improvement of $2.64\%$ and $5.53\%$ accordingly.
\methodname~ achieves consistent performance gain on the metrics.
This further confirms model-agnostic effective generalization of \methodname~ to the unseen environment.

\begin{figure*}[t]
    \centering
    \includegraphics[width=.93\textwidth]{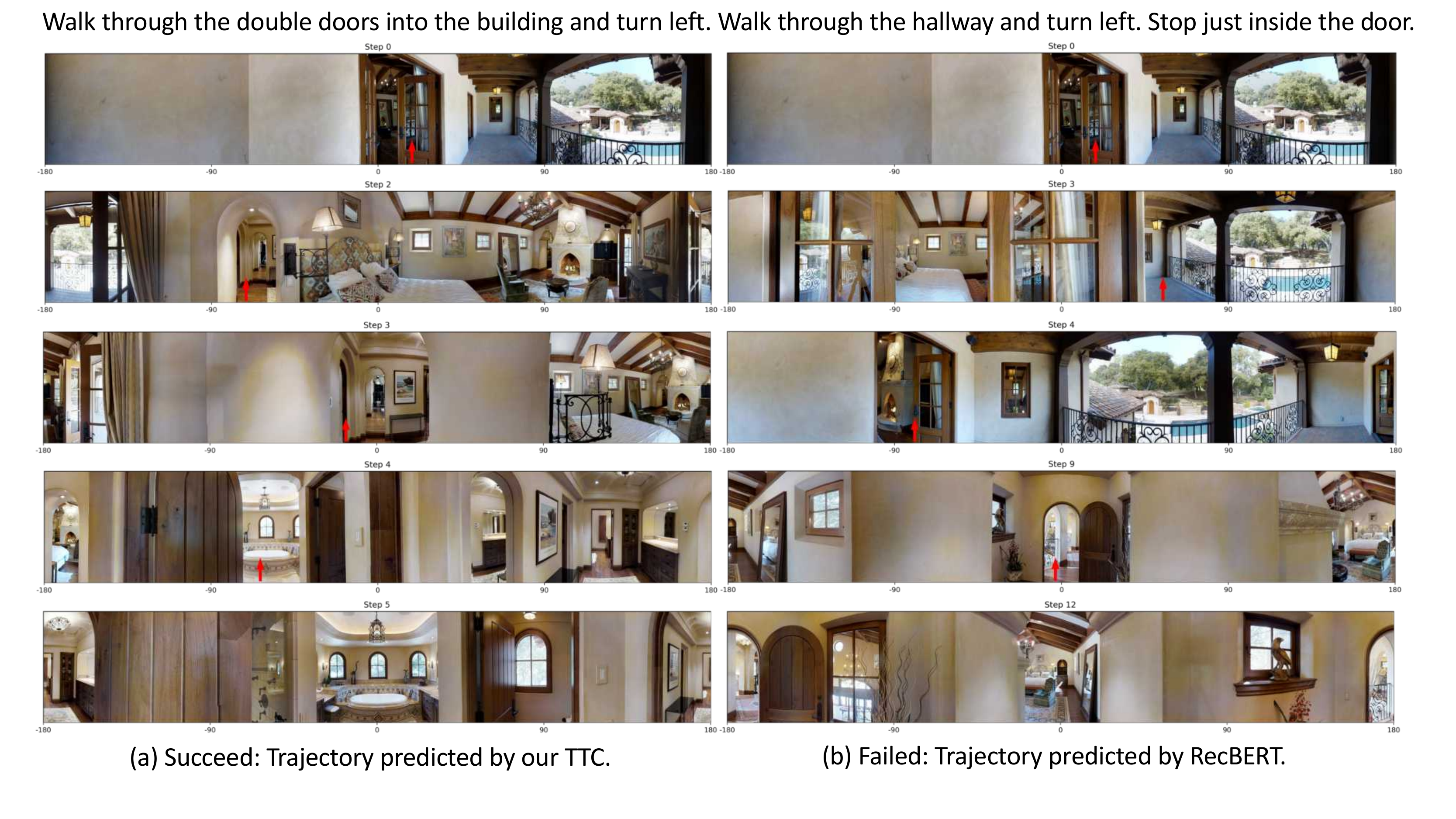}
    \caption{\textbf{Navigation Trajectory Visualization.} The panoramic view of the start point, intermediate steps and the stop point of the predicted trajectories by \bert~ and \methodname~ are visualized.}
    \label{fig:trajectory}
\end{figure*}

\subsection{Module Ablation}
\label{sec:module_ablation}
\noindent \textbf{Effect of Test-time Adaptation}
To study the effect of test-time adaptation (\TTA~), we show results of both the full \methodname~ model and Nearest Neighbor Contrastive (\NNC~) variant without \TTA~.
As shown in Table~\ref{tab:r2r} and Table~\ref{tab:rxr}, the test-time adaptation pushes the performance improvement even higher.
For \clip~, \TTA~ achieve $6.81\%$ and $6.80\%$ improvement over \NNC~ on SR and SPL on Validation Seen, $3.54\%$ and $1.87\%$ on Validation Unseen.
For \bert~, \TTA~ achieve $7.98\%$ and $8.97\%$ improvement over \NNC~ on SR and SPL on Validation Seen, $7.98\%$ and $6.98\%$ on Validation Unseen.
This indicates the test-time adaptation is irreplaceable for learning the generalization to the unseen environment.

\noindent \textbf{Effect of Objectives}
We further study the effect of each component of our proposed framework over the R2R benchmark.
To understand the contribution of each component, we split the objectives into three parts, $ML$, $CL_{IL}$ and $CL_{RL}$.
Results are reported in Table~\ref{tab:r2rablation}.
$ML$ represents the objective of the supervised part of \methodname~, a combination of Imitation Learning (IL) and Reinforcement Learning (RL), which is adopted from the common architecture applied in vision-and-language navigation.
To ensure visual consistency, we have a contrastive learning objective of IL $CL_{IL}$ and a contrastive learning objective of RL $CL_{RL}$.
On Validation unseen, the \clip~ based variant without $CL_{IL}$ objective experience $6.60\%$ and $3.42\%$ drop on SR and SPL, respectively.
The variant without $CL_{RL}$ objective experiences $5.63\%$ and $3.42\%$ drop on SR and SPL, respectively.
We observe $16.62\%$ and $15.81\%$ drop of the \bert~ based variant without $CL_{IL}$ objective on SR and SPL, respectively.
Without $CL_{RL}$ objective experience $13.16\%$ and $16.35\%$ drop on SR and SPL respectively.
% \yj{re-write this paragraph}
To study the performance of a pure self-supervised variant of \methodname~, we remove the $ML$ objective and train the model only with $CL_{IL}$ and ${CL_{RL}}$.
On Validation Seen and Unseen, the self-supervised \methodname~ based on \clip~ and \bert~ experience performance drop, which indicates the importance of the supervised part.
% insight here?

\noindent \textbf{Efficiency Analysis}
We keep the test-time augmentations at a reasonable number and achieve a balance between efficiency and accuracy.
We show the inference time comparison on the R2R benchmark.
For validation unseen split, \clip~ and \bert~ baseline spent $23$ seconds and $85$ seconds.
With test-time adaptation, the inference time increase to $70$ and $190$ seconds.
For validation seen split, \clip~ and \bert~ baseline spent $23$ seconds and $40$ seconds.
With test-time adaptation, the inference time increase to $85$ and $89$ seconds.
The increased time is mainly consumed by the extra adaptation and augmentations.
When applying selective augmentations and test-time adaptation, the inference time is $41$, $34$ for \clip~ based and $57$, $103$ for \bert~ on seen and unseen, respectively.
This indicates the balance between efficiency and accuracy.

\subsection{Qualitative Analysis}
% \subsubsection{Navigation Trajectory}
In Figure~\ref{fig:trajectory}, we visualize the navigation trajectory of baseline and \methodname~ with the architecture of \bert~ under R2R unseen environment.
We show the panoramic view of the start point, intermediate steps, and stop point.
Given the same instruction and the start point view, the visualization of trajectory predicted by both models demonstrates the generalization superiority of our \methodname~.
\bert~ failed to recognize the double door of the building and make wrong action prediction to the next entrance. It made the wrong inference of the hallway and stops at the wrong position.
While our \methodname~ successfully refer to the double doors and follow the instruction inside the building, and turn left at the correct viewpoint. Thus select the right hallway to walk through and turn left. Finally, the agent stops just at the targeted position.
More trajectory visualization and the bird view comparisons can be found in Appendix~\ref{sec:appendix_qualitative}.

\section{Related Work}
\noindent \textbf{Vision-language Navigation}
~\citep{Wang2019reinforcedvln, Tan2019LearningTN, shen2021much, Hong2021VLNBERTAR} learn to ground cross-modal reasoning challenge via imitation learning and reinforcement learning.
Prior studies ~\citep{fu2020counterfactual, Parvaneh2020CounterfactualVN, Liu2021VisionLanguageNW, Majumdar2020ImprovingVN} achieve improvement of the generalizability by proposing diversified forms of augmentations.
Compared with the most related work, ~\citet{wang2019reinforced} adapt model in all the samples, and ~\citet{ray2019adspath} adpat in the house-level.
In contrast, we adopt a more realistic setting that adapt in one single sample without any assumptions of the house information, which further improves the generalization for the VLN agent.
Besides, \methodname~ is model-agnostic to either the sequence-to-sequence or transformer-based architectures.

\noindent \textbf{Test-time Adaptation}
Test-time adaptation aims at leveraging test assumptions to address the problem of distribution shift.
Most of the previous studies can be divided into two categories, test-time training and test-time augmentation.
Test-time training (TTT) ~\citep{Sun2020TestTimeTW} proposes a model that composes of both the supervised module and self-supervised module and enables a different training procedure on test inputs.
Test-time augmentation (TTA) has been proved effective in addressing the problem of distribution shift and achieving generalization to test set.
We propose to combine the advantages of both TTT and TTA regime with a two-stage semi-supervised model design with considerations of maximizing agreements over augmentations to adapt to distribution shift at test-time.

\section{Conclusion}
We propose a Test-time Visual Consistency (\methodname~) framework that improves the generalizability of the VLN agents to unseen environments.
\methodname~ is composed of a supervised branch and a self-supervised branch based on Imitation Learning and Reinforcement Learning.
In addition to standard semi-supervised joint training procedure, \methodname~ implements test-time adaptation before inference.
Experimental results on R2R and RxR benchmarks confirm the superiority of \methodname~ over previous state-of-the-art VLN baselines.

\section*{Limitations}
In general, the limitations of our work are from three aspects.
First, the visual backgrounds in the dataset we are using are from the English-Speaking country.
Though the multi-lingual instructions are provided, they are still using the English instructions as pivot and thus biased to the English language navigation habits.
Second, we focus on the in-door environment with sufficient data and may result in limited performance in the low-resource situation.
Finally, how to scale up the framework to multi domain settings remain underexplored.
In the future work, we hope to solve the issues together in a simple unified setting and bring the model to broader VLN applications.

\section*{Ethics Statement}
\label{sec:ethical}
Since the VLN challenge is a fundamental problem in the field of vision and language, we do not foresee any significant ethical issues. Minor concerns are the biases introduced from the experimental dataset and baseline model side. In fact, the dataset we used (R2R, RxR), and the baseline model we used (see \cref{sec:exp_setup}) are all published. Thus these concerns seem low-risk for our experimental evaluations. In addition, we did not notice any such problems in our work.

% \section*{Acknowledgements}
% This document has been adapted by Yue Zhang, Ryan Cotterell and Lea Frermann from the style files used for earlier ACL and NAACL proceedings, including those for 
% ACL 2020 by Steven Bethard, Ryan Cotterell and Rui Yan,
% ACL 2019 by Douwe Kiela and Ivan Vuli\'{c},
% NAACL 2019 by Stephanie Lukin and Alla Roskovskaya, 
% ACL 2018 by Shay Cohen, Kevin Gimpel, and Wei Lu, 
% NAACL 2018 by Margaret Mitchell and Stephanie Lukin,
% Bib\TeX{} suggestions for (NA)ACL 2017/2018 from Jason Eisner,
% ACL 2017 by Dan Gildea and Min-Yen Kan, NAACL 2017 by Margaret Mitchell, 
% ACL 2012 by Maggie Li and Michael White, 
% ACL 2010 by Jing-Shin Chang and Philipp Koehn, 
% ACL 2008 by Johanna D. Moore, Simone Teufel, James Allan, and Sadaoki Furui, 
% ACL 2005 by Hwee Tou Ng and Kemal Oflazer, 
% ACL 2002 by Eugene Charniak and Dekang Lin, 
% and earlier ACL and EACL formats written by several people, including
% John Chen, Henry S. Thompson and Donald Walker.
% Additional elements were taken from the formatting instructions of the \emph{International Joint Conference on Artificial Intelligence} and the \emph{Conference on Computer Vision and Pattern Recognition}.

\bibliography{anthology,custom}
\bibliographystyle{acl_natbib}

\newpage
\appendix

\section{Appendix}
\label{sec:appendix}
\subsection{Method Details}
\label{sec:method_details}
\begin{figure*}[t]
    \centering
    \includegraphics[width=.9\textwidth]{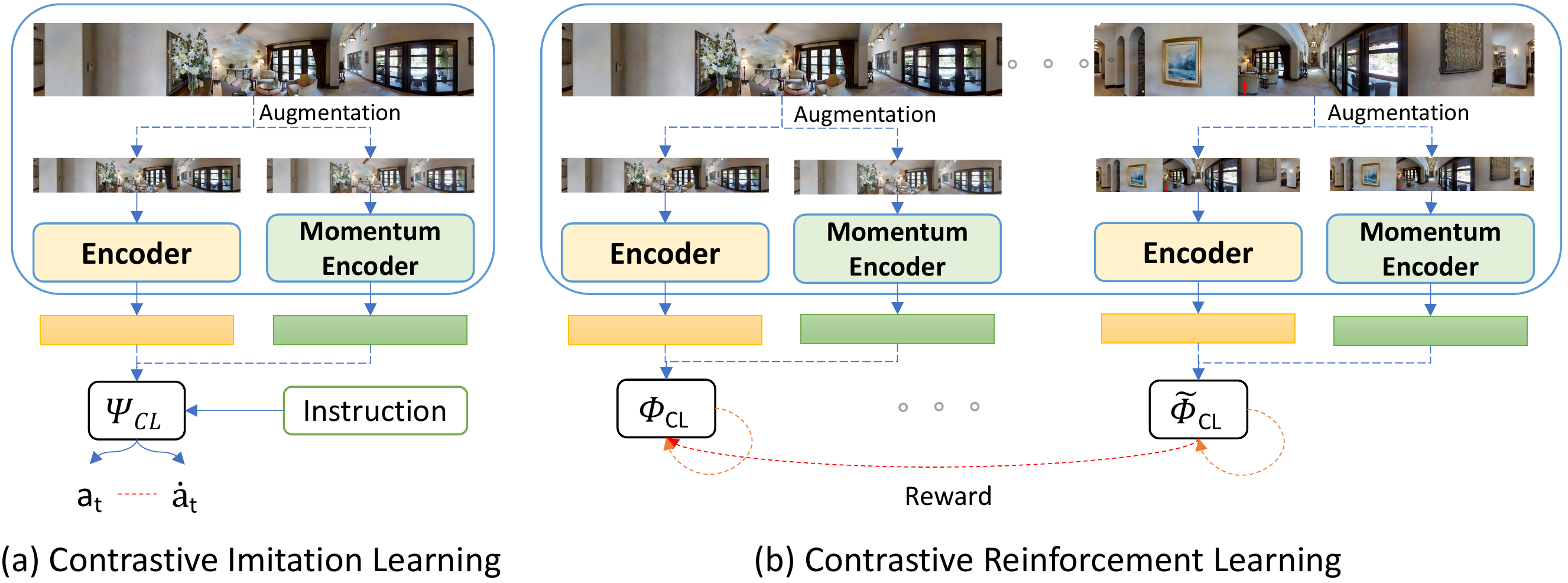}
    \caption{\textbf{Contrastive Module.} (a) The policy $\psi$ learns by attracting positive samples in the predicted action embedding space. (b) The policy $\phi$ regularizes the reward for being similar across positive samples ($a_{t}$ and $\dot{a}_{t}$) for RL. The visual consistency is encouraged between visual embedding of Encoder and Momentum Encoder.}
    \label{fig:vlnContrastive}
\end{figure*}

As shown in Figure~\ref{fig:vlnContrastive}(a), the policy $\psi_{CL}$ network learn to maximize the agreement of action decisions $a_t$ and $\hat{a}_t$ over positive observation pairs.
As shown in Figure~\ref{fig:vlnContrastive}(b), we consider taking augmentations for introducing consistency during actor critic learning.

\begin{figure*}[t]
    \centering
    \includegraphics[width=0.9\textwidth]{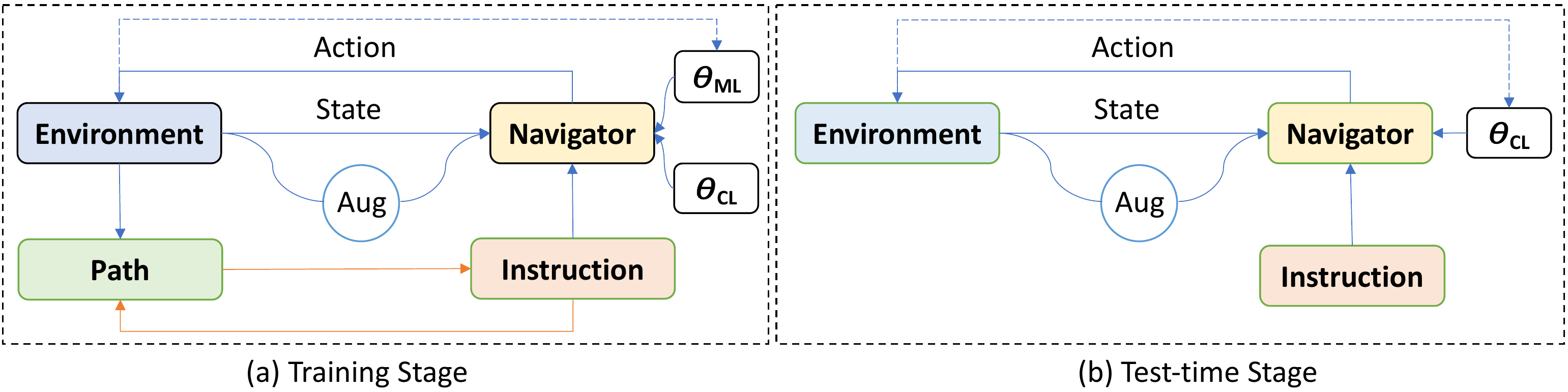}
    \caption{\textbf{A Separate View of Training and Test-time Procedure.} For the training stage, the $\theta_{ML}$ and $\theta_{CL}$ are updated jointly with Equation~\ref{eq:obtrain}. At test-time, the $\theta_{ML}$ is fixed, and the $\theta_{CL}$ is adapted using Equation~\ref{eq:obtest}.
    After the adaptation, the action distribution is predicted by the combination of $\theta_{ML}$ and $\theta_{CL}$.
    }
    \label{fig:procedure_all}
\end{figure*}

\subsection{Background}
\label{sec:background}
\noindent \textbf{Vision-language Navigation}
~\citep{fu2020counterfactual, Parvaneh2020CounterfactualVN} utilize the counterfactuals to augment either the trajectories or the visual observations.
~\citep{Liu2021VisionLanguageNW} proposes to mixup the environment.
~\citep{Majumdar2020ImprovingVN} leverages large-scale image-text pairs for the Web Data.
By training on a large amount of image-text-action triplets in a self-supervised learning manner, the pre-trained model provides generic representations of visual environments and language instructions.
Some studies ~\citep{hu2019looking, gan2020look} learn to utilize other available modalities.
~\citep{hao2020learning} presents pre-training and fine-tuning paradigm for vision-and-language navigation (VLN) tasks.
~\citep{wang2019reinforced} propose a novel Reinforced Cross-Modal Matching (RCM) approach that enforces cross-modal grounding both locally and globally via reinforcement learning (RL) to address three critical challenges for this task: the cross-modal grounding, the ill-posed feedback, and the generalization problems.

\noindent \textbf{Consistency Regularization}
~\citep{Xie2021PropagateYE, melaskyriazi2021pixmatch} learn dense feature representations via pixel-level visual consistency.
~\citep{Ouali2020SemiSupervisedSS, Abuduweili2021AdaptiveCR, Jeong2021FewshotOR} learn consistency regularization for semantic segmentation.
% For example, ~\citep{Ouali2020SemiSupervisedSS} propose to maintain the context-aware consistency to make robust representations for varying environments.~\citep{Ren2021AdaptiveCP} proposes the Adaptive Consistency Prior for Image Denoising.
% These studies expand the area of domain adaptation, transfer learning and few shot learning.
% ~\citep{Zhang_2021_CVPR} proposes to enforce the temporal consistency in low light video enhancement with static images.
~\citep{Ren2021AdaptiveCP} proposes an Adaptive Consistency Prior based Deep Network for Image Denoising.
Recent studies~\citep{Li20213DHA, Hyun2021SelfSupervisedVG, Yan2021SelfAlignedVD} show increasing interest of encouraging consistency across multiple modalities.
~\citep{Srinivas2020CURLCU} learns contrastive unsupervised representations for reinforcement learning.
~\citep{Hippocampus2020ASI} proposes to train the policy as a classifier via contrastive regularization for imitation learning.
In this study, we propose to leverage consistency into both imitation learning and reinforcement learning process of the VLN agent. 
Specifically, we utilize momentum contrast for enforcing visual consistency between query panoramic features as well as stable decision makings across similar observations.

\subsection{Dataset Details}
\label{sec:dataset_details}
The details of our experimental datasets are:
\begin{itemize}
    \item {\verb|R2R|~\citep{Anderson2018VisionandLanguageNI}}: is a VLN dataset collected in photo-realistic environments (Matterport3D~\citep{Chang2017Matterport3DLF}). It contains $61$, $11$ and $18$ scenes for training, validation and testing, respectively.
    \item {\verb|RxR|~\citep{ku2020room}}: is a multilingual (English, Hindi, and Telugu) and larger (more extended instructions and trajectories) than other existing VLN datasets. It contains $11089$, $232$, $1517$ and $2684$ paths in train, val-seen (train environments), val-unseen (val environments), and test splits respectively.
\end{itemize}

\subsection{Baseline Details}
\label{sec:baseline_details}
We compare the single-run performance of different agents including Seq2Seq~\citep{Anderson2018VisionandLanguageNI}, Speaker-Follower~\citep{fried2018speakerfollower}, PRESS~\citep{Li2019press}, AuxRN~\citep{Zhu2020AuxRN}, PREVALENT~\citep{Hao2020prevalent} and RelGraph~\citep{Hong2020relgraph} on R2R, EnvDrop~\citep{Tan2019LearningTN} and Syntax~\citep{Li2021syntax} on RxR.
We conduct experiments over both benchmarks with our model-agnostic \methodname~ added to baselines: 1) CLIP-VIL~\citep{shen2021much} is an extension of EnvDrop~\citep{Tan2019LearningTN}, which leverages the representation from large-pretrianed CLIP~\citep{Radford2021LearningTV} model to replace the visual features extracting backbone from ResNet~\citep{He2016DeepRL} to Vision Transformer (ViT~\citep{Dosovitskiy2021AnII}). 2) VLNBERT~\citep{Hong2021VLNBERTAR} proposes a recurrent BERT model that is time-aware for use in VLN. The parameters of our implemented baseline are initialized from PREVALENT~\citep{Hao2020prevalent}.
\begin{itemize}
    \item {\verb|CLIP-VIL|~\citep{shen2021much}}: is an extension of EnvDrop~\citep{Tan2019LearningTN}, which leverages the representation from large-pretrianed CLIP~\citep{Radford2021LearningTV} model to replace the visual features extracting backbone from ResNet~\citep{He2016DeepRL} to Vision Transformer (ViT~\citep{Dosovitskiy2021AnII}).
    \item {\verb|VLNBERT|~\citep{Hong2021VLNBERTAR}}: proposes a recurrent BERT model that is time-aware for use in VLN. 
    The parameters of our implemented baseline are initialized from PREVALENT~\citep{Hao2020prevalent}.
\end{itemize}

\begin{figure*}[t]
    \centering
    \includegraphics[width=\textwidth]{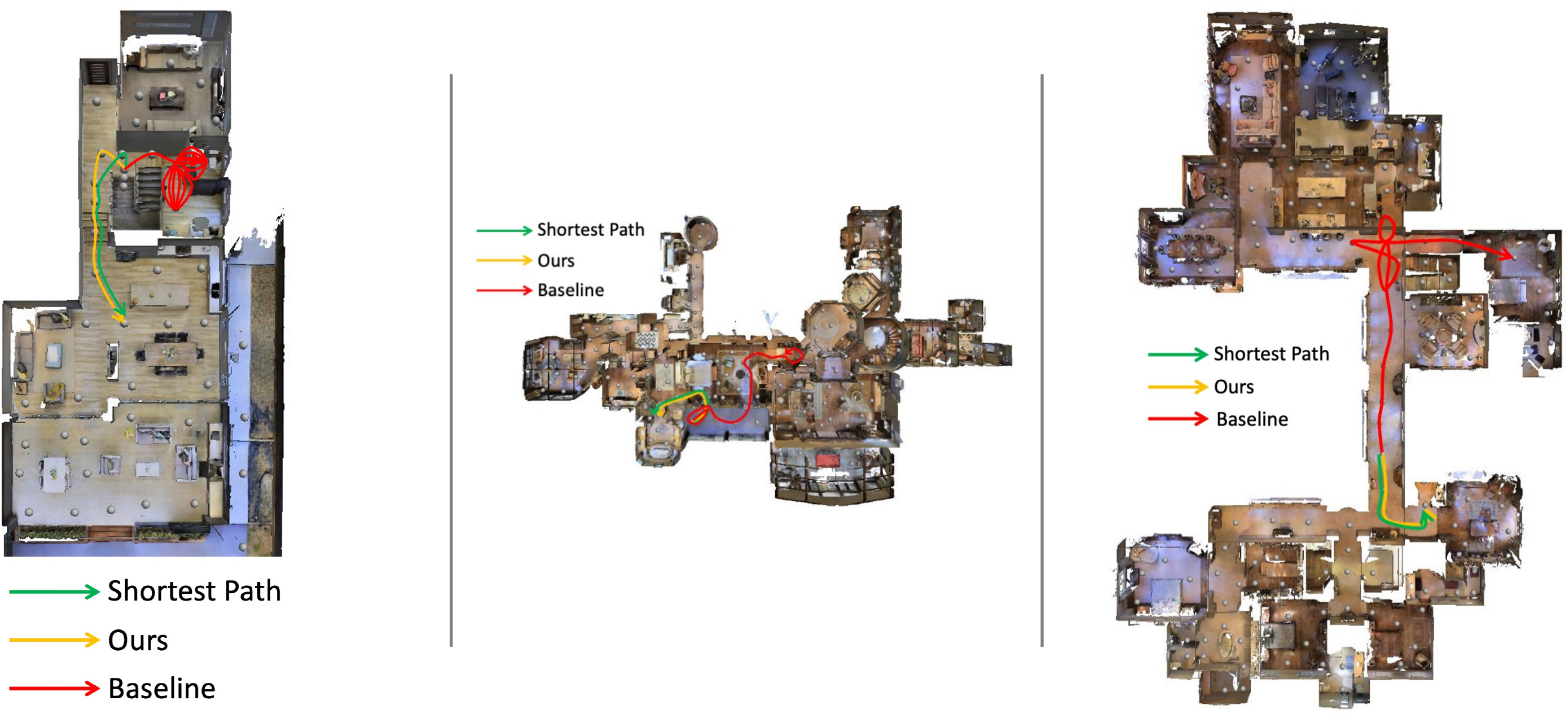}
    \vspace{-2ex}
    \caption{\textbf{Bird View Comparison.} The green line represents the trajectory of ground truth (Shortest Path). The red line and the yellow line represent the trajectory of \bert~ baseline and \bert~ based \methodname~ respectively. The trajectory points from the start position to the end position. The leftmost is Scene A, and the rightmost two scenes are Scene B.} 
    \vspace{-3ex}
    \label{fig:birdview}
\end{figure*}

\begin{figure*}[t]
    \centering
    \includegraphics[width=\textwidth]{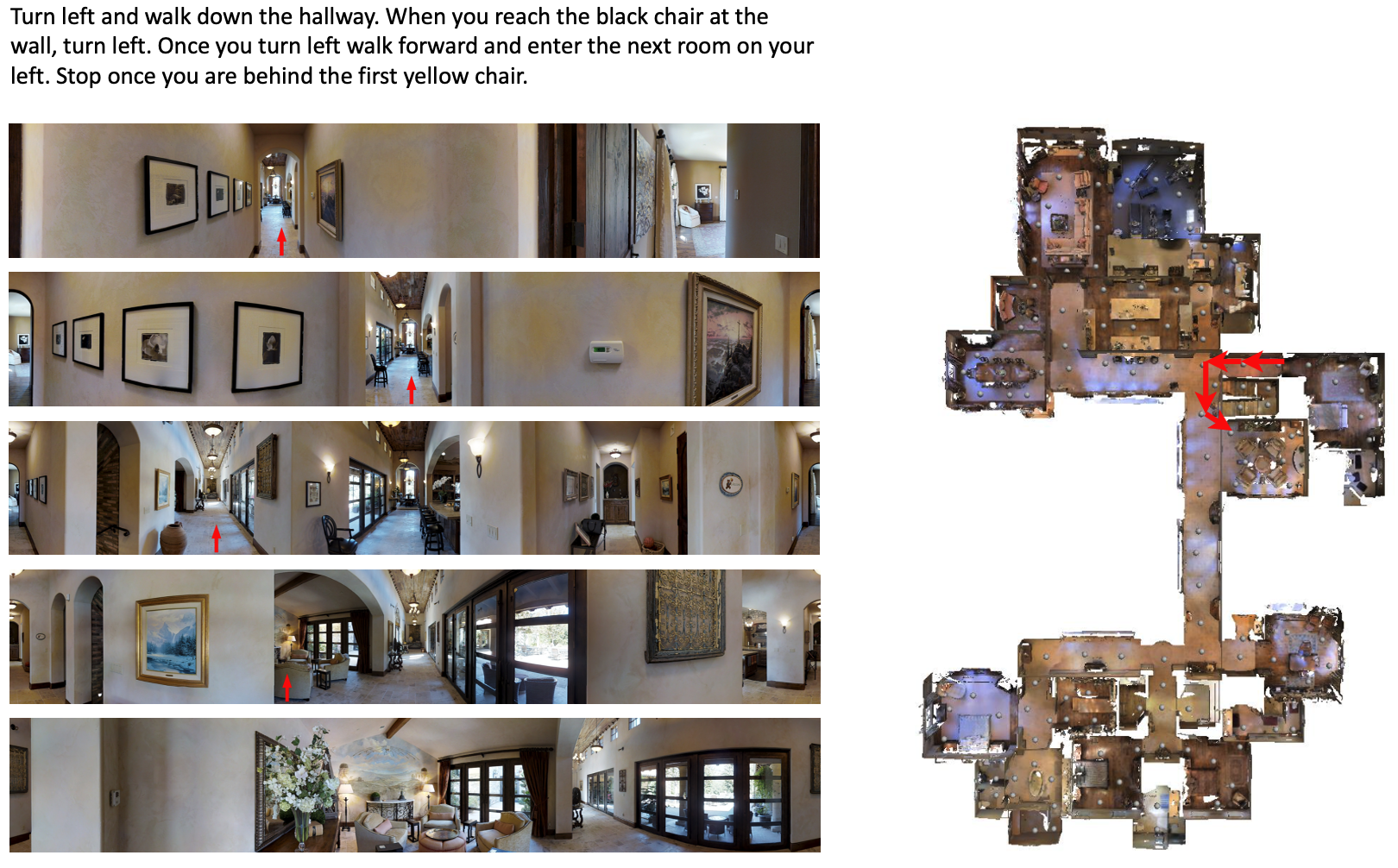}
    \caption{Trajectory Visualization.}
    \label{fig:birdview_trajectory}
\end{figure*}

\subsection{Implementation Details}
~\label{sec:implementation_details}
There are $K= 36$ view images in each panoramic observation with available panoramic action space.
The observation features are ResNet image features for \bert~ and ViT image features for ~\clip~.
The language encoder is LSTM for sequence-to-sequence architecture and BERT for transformer architecture.
The architecture consists of a Speaker and a Listener following the implementation in EnvDrop.
The speaker is either a sequence-to-sequence or a transformer-based module to estimate the likelihood of instruction for the given trajectory.
The listener accordingly follows the instruction from the speaker that estimates the likelihood of action sequence for the instruction-trajectory pair.
We use Adam Optimizer and batch sizes $32$ for training and test-time training.
The learning rates are set as $1e-4$.
We train $20000$ iterations for the training stage and $10$ iterations for the test-time stage.
We set the balance factor $\lambda_{ml}$, $\lambda_{rl}$, $\lambda_{cl_ml}$, $\lambda_{cl_rl}$ as $0.2$, $0.2$, $0.2$ and $0.2$ by grid search respectively.
The parameter size of our \methodname~ based on \clip~ and \bert~ is $91$M and $160$M respectively.

\noindent \textbf{Computation and Resources}
We use four single NVIDIA A100 GPU Server for all the experiments.
The computation is used for the training and testing stage.
The average runtime for our proposed \methodname~ is discussed in Section~\ref{sec:module_ablation}.

\subsection{Qualitative Results}
\label{sec:appendix_qualitative}
To further look at the high-level decision-making process of the agent, we visualize the bird view of trajectories predicted by the baseline and \methodname~, compared with ground truth trajectory in Figure~\ref{fig:birdview}.
We randomly select one case for Scene A and two cases for Scene B under the R2R validation unseen environment.
For Case 1 under Scene A, the trajectory predicted by \methodname~ matches the Shortest Path.
Meanwhile, the \bert~ baseline heads the wrong way initially and ends up with redundant trajectories.
Similar to Case 2 and Case 3 under Scene B, the baseline is more likely to miss the key reference in the new environment and head in the wrong direction.
In contrast, our proposed \methodname~ shows incredible generalization under unseen environments and follows the instruction continually.
We also showcase a trajectory visualization in Figure~\ref{fig:birdview_trajectory}.

\end{document}